\documentclass[conference]{IEEEtran}
\IEEEoverridecommandlockouts

\usepackage{graphicx}
\usepackage{algorithmic}
\usepackage{graphicx}
\usepackage{textcomp}
\usepackage{cite}
\usepackage{amsmath,amssymb,amsfonts}
\usepackage{algorithmic}
\usepackage{graphicx}
\usepackage{textcomp}
\usepackage{xcolor}
\usepackage{soul}
\usepackage{balance}
\usepackage{colortbl}
\usepackage{multirow}
\usepackage{float}
\usepackage{amsmath}
\usepackage{caption}
\usepackage{subcaption}
\usepackage{float}
\usepackage{chngpage}
\usepackage{tcolorbox}
\usepackage[ruled,vlined]{algorithm2e}
\usepackage{tikz}
\usepackage[hyphens]{url}
\usepackage{hyperref}
\def\BibTeX{{\rm B\kern-.05em{\sc i\kern-.025em b}\kern-.08em
    T\kern-.1667em\lower.7ex\hbox{E}\kern-.125emX}}

\begin{document}

\title{On Assessing The Safety of Reinforcement Learning algorithms Using Formal Methods}

\author{\IEEEauthorblockN{Paulina Stevia Nouwou Mindom, Amin Nikanjam, Foutse Khomh, John Mullins}
\IEEEauthorblockA{\textit{SWAT Lab., Polytechnique Montréal}\\ 
\textit{Montréal, Canada}\\
\{paulina-stevia.nouwou-mindom, amin.nikanjam, foutse.khomh, john.mullins\}@polymtl.ca}
}

\maketitle

\begin{abstract}
The increasing adoption of Reinforcement Learning in safety-critical systems domains such as autonomous vehicles, health, and aviation raises the need for ensuring their safety. Existing safety mechanisms such as adversarial training, adversarial detection, and robust learning are not always adapted to all disturbances in which the agent is deployed. Those disturbances include moving adversaries whose behavior can be unpredictable by the agent, and as a matter of fact harmful to its learning. Ensuring the safety of critical systems also requires methods that give formal guarantees on the behaviour of the agent evolving in a perturbed environment. It is therefore necessary to propose new solutions adapted to the learning challenges faced by the agent. In this paper, first we generate adversarial agents that exhibit flaws in the agent's policy by presenting moving adversaries. Secondly, We use reward shaping and a modified Q-learning algorithm as defense mechanisms to improve the agent's policy when facing adversarial perturbations. Finally, probabilistic model checking is employed to evaluate the effectiveness of both mechanisms. We have conducted experiments on a discrete grid world with a single agent facing non-learning and learning adversaries. Our results show a diminution in the number of collisions between the agent and the adversaries. Probabilistic model checking provides lower and upper probabilistic bounds regarding the agent's safety in the adversarial environment.
\end{abstract}


\section{Introduction} \label{section1}

Machine Learning (ML) has gained tremendous interest in the academic and industrial fields, due to the increased data volumes, advanced algorithms, and improvements in computing power and storage \cite{l2017machine}. Among ML algorithms, Reinforcement Learning (RL) has been positively recognized for the ability of agents to learn through their interactions with the environment in which they are deployed. This technique has been useful to collect informations about the environment of some applications \cite{sallab2017deep, kormushev2013reinforcement}. 
On the downside, interacting with the environment gives rise to some safety challenges, linked to the level of criticality of the applications/domains involved.
 
The safety of RL agents has been studied in the literature. Some approaches consist of retraining the model with adversarial examples to improve the agent's resilience to them \cite{kos2017delving,pattanaik2018robust}. 
Some other approaches consist of designing a controller to find an optimal policy for the agent in presence of disturbances \cite{li2018training, morimoto2005robust}. Improving the safety of RL agents has also been studied in the context of adversaries represented by fixed (non-moving) obstacles. In \cite{singla2019memory, cimurs2020goal}, the agent was trained to avoid obstacles existing in the environment. Collision avoidance applications have been studied in the literature \cite{sangiovanni2018deep, theie2020deep} but to the best of our knowledge, no research has been conducted for moving obstacles which we referred to as moving adversaries. Moreover, providing formal guarantees on the safety of an agent evolving in a perturbed environment has not receive enough attention in the literature. 
This is due to the uncertainty of the environment which makes the problem very challenging \cite{bouton2019reinforcement}. 

In this paper, we aim to assess the safety of a RL agent against moving adversaries and give provable guarantees.The agent is trained through an online RL algorithm in presence of moving adversaries. At the end of the training, the trained agent appears to show some weaknesses linked to the vulnerability during the training. We first consider two types of moving adversaries that uncover some flaws in the learnt policy of the agent: static adversary (non-adaptive and non-learnable) and learnable adversary. Actually, these adversaries create threats that prevent the agent from achieving its objective. Then, we have implemented two defense mechanisms to improve the learning process. One of them is based on reward shaping, a method for engineering the reward function to guide the agent during training. The second defense mechanism consists of modifying the RL algorithm by instructing the agent to avoid the adversary during training. Finally, we provide formal probabilistic guarantees regarding both defense mechanisms. Experimentally, we have verified reachability properties, i.e., the ability of the agent to reach specific states of the environment \cite{baier2008principles} on the well-known grid world environment to examine the safety and overall performance. 
The results show that by applying the proposed defense mechanisms, the agent is capable of performing more safely in the presence of various adversaries, e.g., by reducing the collision rate. Formal guarantees are provided to assess the effectiveness of our defense mechanisms, by computing probabilities of our reachability properties. Our key findings in this paper are:
\begin{itemize}
\item An adversarial environment consisting of static/learnable moving adversaries can badly harm the agent policy,
\item Strengthening the agent policy with our proposed defense mechanisms helps in improving the agent's behaviour, 
\item Probabilistic model checking provides a range of guarantees on the safety of the agent's behaviour.
\end{itemize}
The rest of this paper is organized as follows. Section \ref{section2} briefly reviews related works on the safety of RL agents. In Section \ref{section3}, we present the required background knowledge on RL, reward Shaping, and probabilistic model checking. Section \ref{section4} describes our methodology and Section \ref{section5} presents the results of our experiments. Finally, we present concluding remarks in Section \ref{section6}.

\section{Background}\label{section3}
\subsection{Reinforcement Learning}
\begin{figure}
        \centering
        \includegraphics[width=0.4\textwidth]{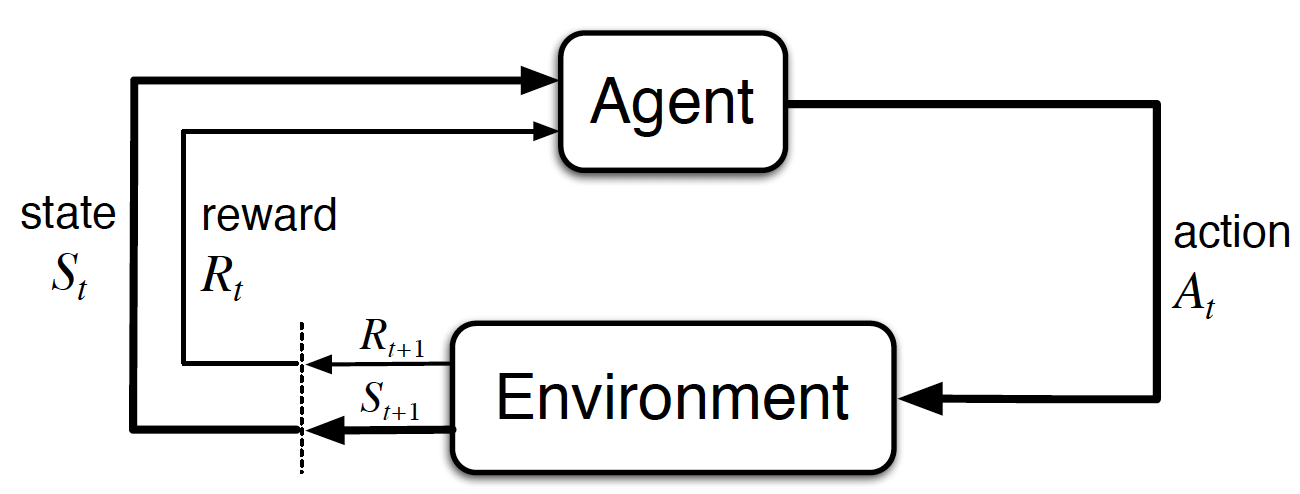}
        \caption{Interaction between the agent and the environment in a Markov decision process~\cite{sutton2018reinforcement}.}
        \label{MDP:Figure}
    \end{figure}

A RL problem can be mathematically described as an MDP \cite{sutton2018reinforcement}. Figure \ref{MDP:Figure} illustrates the interaction of the agent with its environment in a MDP. An MDP is defined by:
\begin{itemize}
    \item A set of states S,
    \item A set of actions A,
    \item A transition function determining the probability that the agent, moves to the state $s'$ and gets $r$ as a reward given the preceding state $s$ by taking action $a$:
    \begin{equation} \label{}
  p(s', r|s, a)= Pr \{S_{t}=s',R_{t}=r | S_{t-1}=s,A_{t-1}=a\} 
\end{equation}
    \item The reward function $R_{t}(s,a,s')$ representing the result of the agent’s action,
    \item The starting state $s_0$,
    \item The discount factor $\gamma $ which expresses how much a reward counts \cite{petrik2009biasing}.
\end{itemize}
The beginning of a trajectory in an MDP can be written as follows:
\begin{equation}
    S_{0},~ A_{0},~ R_{1},~S_{1},~ A_{1},~ R_{2},~S_{2},~ A_{2},~ R_{3}... 
\end{equation}
The trajectory describes the succession of states, actions taken by the agent and the rewards it received over time. The agent (also called the \textit{learner}) interacts with the environment, takes actions allowing him to evolve in different states of the environment. The latter sends back feedback in the forms of rewards, values that the agent attempts to maximize over time.
 
The agent’s objective is to find the optimal policy that maximizes the cumulative expected reward discounted which can be done using Bellman equations. They express how good it is for an agent to be in a state $s$ following the policy $\pi$: Equation \ref{vs:equation}. 
\begin{equation}
  V^{\pi}(s)= \underset{a}{max} \bigg( \sum   _{s',r} p(s',r|s,a)[r+\gamma V^{\pi}(s') ] \bigg)
  \label{vs:equation}
\end{equation}
 
Q-learning algorithm \cite{dayan1992q} is a well-known RL algorithm to train the agent when the model of the environment is unknown. In Q-learning, the action-value function tells us how good it is to take a particular action from a particular state following the policy $\pi$; Equation \ref{qs:equation}.
 
\begin{equation}
     Q^{\pi}(s,a)=\bigg( \sum   _{s',r} p(s',r|s,a)[r+\gamma \underset{a'}{max}  Q^{\pi}(s',a') ] \bigg)
     \label{qs:equation}
\end{equation}

At the $n^{th}$ step of an episode, the agent:
\begin{itemize}
    \item Observes its current state $s$
    \item Selects and performs an action $a$
    \item Observes the next state
    \item Receives an immediate reward $r$
    \item Adjusts its Q-values by using the learning rate $\alpha$ and the discount factor $\gamma$
    \begin{equation}
         Q^{\pi}(s,a)= Q^{\pi}(s,a)+ \alpha(r + \gamma  \underset{a'}{max}Q^{\pi}(s',a')-
   Q^{\pi}(s,a))
    \end{equation}
\end{itemize}
 
The agent has to decide between exploration and exploitation \cite{poole2010artificial}. The agent can explore the environment to have a better estimation of the Q-values or can exploit its knowledge and always choose the action that maximizes the Q-values. To trade-off between exploration and exploitation, one can implement the $\epsilon$-greedy algorithm ($ 0 \leq \epsilon \leq 1$) to select a random action $\epsilon$ time and the action that maximize $Q^{\pi}(s,a)$, ~ $1-\epsilon$ time. It is possible to change $\epsilon$ through time so that in the beginning the agent should select actions randomly, as soon as the environment is explored it should act more greedily.
 
\subsection{Reward Shaping}
Reward shaping consists of modifying the original reward function with a shaping function that incorporates domain knowledge \cite{hu2020learning}. 
The new reward function can be formalized as in Equation \ref{r:equation}:
\begin{equation}
  r'= r + F
  \label{r:equation}
\end{equation}
where $r$ is the original reward function, $F$ is the shaping reward function, and $r'$ is the modified reward function. Potential-based reward shaping (PBRS) \cite{ng1999policy} is the first approach that guarantees invariance of the optimal policy. Specifically, PBRS defines $F$ as the difference of potential values:
\begin{equation}
     F (s, a, s')= \gamma \phi(s') - \phi (s)
     \label{f:equation}
\end{equation}
where $ \phi  : S \xrightarrow[]{} \mathbb{R} $ is a potential function which gives some kind of hints on states \cite{hu2020learning}.
 
Ng et al. \cite{ng1999policy} demonstrates that every optimal policy produced under the new reward shaping function will also be an optimal policy without the shaping process.
 
\subsection{Probabilistic Model Checking}
Probabilistic model checking is a formal verification process of systems with stochastic behaviour. It allows us to analyse quantitative properties on those systems. PRISM \cite{hinton2006prism} is the tool that we use to verify the satisfiability of the RL agent properties. It is a probability model checker that takes as input a probabilistic model, a temporal logic property and returns whether or not the property is satisfied on the model.  We have extracted a DTMC (Discrete-time Markov chain) from the agent policy, whose behaviour at each time is described as a discrete probabilistic choice over several outcomes. A DTMC is a tuple $D=(S,s_i,P,AP,L)$ \cite{kwiatkowska2010advances} where :
\begin{itemize}
    \item $S$ is a set of states,
     \item $s_i \in S$ is an initial state,
     \item $P : S ~ \mathrm{X} ~ S \rightarrow [0, 1]$ ~ is a transition probability matrix such that ~ $\sum_{s^{\prime} \in S}  P(s,s^{\prime}) = 1$ for all $s \in S$,
     \item $AP$ is a set of atomic propositions,
     \item $L : S \to 2^{AP}$ is a labelling function that assigns, to
each state $s \in S$, a set $L(s)$ of atomic propositions.
\end{itemize}
PCTL (Probabilistic Computation Tree Logic) \cite{hansson1994logic} specifications can be verified on DTMC. Their formalism is as follow:
\begin{itemize}
    \item $\Phi ::= true ~|~ a ~|~ \neg ~ \Phi~|~ \Phi~ \wedge~ \Phi~ |~ \Phi ~\vee~ \Phi ~|~ P_{\sim p} [\phi]$
    \item $\phi ::= X\Phi ~|~  \Phi_1 ~ U^{\le k} ~ \Phi_2 ~ |~\Phi_1 ~ U ~ \Phi_2 $
\end{itemize}
where $a$ is an atomic proposition, $\sim  \in $  \{$<, \le,\ge, >$\}, $p \in [0,1]$ and $k ~ \in ~  \mathbb{N} $. A state $s$ satisfies a PCTL formula $\Phi$ if it is true for $s$. $P_{\sim p} [\phi]$ means that the probability of a path formula $\phi$ satisfies the bound $\sim p$. 
For a DTMC path $s_1~ s_2~ s_3~ . . .$, the next state formula $ X \Phi$ holds iff $\Phi$ is satisfied in the next path state (i.e., in state $s_2$); the bounded until formula $\Phi_1 ~ U^{\le k} ~ \Phi_2$ holds iff before $\Phi_2$ becomes true in some state $s_x,~ x < k,~ \Phi_1$ is true for states $s_1$ to $s_{x-1}$; and the unbounded until formula $\Phi_1 ~ U ~ \Phi_2$ removes the constraint $x < k$ from the bounded until formula.

\section{Approach} \label{section4}
\subsection{Environment}
First, we consider an environment that consists of a set of locations. An agent and an adversary are in the environment residing in their locations and can move around. The agent has an objective to reach a specific location which is called the goal. In real-world environments, e.g., autonomous driving, agents can perceive their environment via their sensors, i.e., observations. In this paper, we define observations of agents as a set of the environment's locations: at each time step the agent can observe its current location and the content of its neighboring locations. As such, the agent can observe the adversary when it is in its neighboring locations. The location of the goal is fixed and announced to the agent before it starts its experience in the environment. At each time step: 1) the agent receives its observations from the environment, 2) it selects an action to be taken in the environment (i.e., moving to a neighboring location) and 3) the environment returns a reward signal to the agent. The agent employs a RL algorithm to learn the optimal policy for reaching the goal in the environment. Without loss of generality, we assume that the agent employs the Q-learning algorithm to learn a path (a series of locations to move in consequently) to achieve its objective.  As shown in Algorithm \ref{QObs:algorithm}, the Q-learning algorithm is modified to take into account the observations in order to preemptively avoid the adversary at the next move. We keep track of a Q function, \textbf{$Q_{obs}$}, that we update when the agent has the adversary in its surroundings. In such a case, the Q-value associated to that cell where the adversary is located, is negatively rewarded \textbf{$adv=-100$}. This new Q function is employed whenever the adversary is seen. Otherwise, the normal Q learning algorithm is employed.
 
\subsection{Adversaries}
We design various adversarial policies to test the agent against them; assessing safety and overall performance: non-learnable and learnable adversaries. Both adversaries can move in the environment, the property that makes them different from typical non-moving obstacles. The non-learnable adversaries are designed to patrol in the environment around the goal location to prevent the agent from achieving its objective. Their policies are deterministic and can harm the agent when they are in the same location which is called a collision. It is a threat against the safety of the agent since the ideal path to the goal for the agent should be collison-free. We base the formulation of the agent's safety properties with reference to this ideal path. A safety property describes the consequences of the behavior of the agent for taking the ideal path. During the formal verification process, we will search for collision-free paths to return whether the agent's behavior is safe. The learnable adversary is designed to be smarter, with non-predefined behaviour and the capability of adapting itself to the agent's behaviour as the agent learns how to behave. The learning adversary also uses the Q-learning algorithm. 
 
\subsection{Defense mechanisms}
The adversaries expose flaws in the behaviour of the agent leaving us to design some defense mechanisms to help the agent. We have proposed two defense mechanisms in this paper. The first one is based on PBRS and similar to the work presented in \cite{ng1999policy}, our potential functions are as follows:
\begin{itemize}
    \item The Manhattan distance between the location of the agent and the goal's location.
     \begin{equation}
     \phi (s)= Manhattan(s,g)
     \label{phisM:equation}
\end{equation}
where $Manhattan$ is the Manhattan distance, $s$ the current location of the agent and $g$ the goal's location.
    \item  An extra reward for the agent discounted by the probability of observing no collisions at its current location.
    
    \begin{equation}
     \phi (s)= r * p_c(s)
     \label{phis:equation}
\end{equation}
where $r$ is the original reward function obtained by the agent at location $s$ and $p_c(s)$ is the probability of having no collisions at location $s$. This probability is computed as the number of moves made in a location without colliding with the adversary divided by the total number of moves made to that same location. 
 
\end{itemize}
 
In addition to PBRS, we implement another defense mechanism that is more conservative. The Q-learning algorithm is adopted as shown in Algorithm \ref{Qmodif:algorithm} to instruct the agent to avoid any collisions with the adversary. To do so, we slightly modify the algorithm in its exploitation phase. We check for the presence of the adversary in the neighbouring cells of the agent. When the selected action that has the best q-value leads to a collision with the adversary, we ignore that action and consider the action with the second best q-value.

\begin{algorithm}[t]
\SetAlgoLined
\KwResult{act}

  \eIf{random.uniform(0, 1) $<$ epsilon (the exploration probability)}
  {
   Explore: select a random action \textbf{act}\;
   
   }
   {
   Exploit: select the action \textbf{act} with the max q-value \;
    \If{\textbf{act} leads to adversary}{
    $Q^{\pi}_{obs}(s,act)= Q^{\pi}_{obs}(s,act)+ \alpha( 
    $\textbf{adv}$ + \gamma \underset{a'}{max}Q^{\pi}_{obs}(s',a')- Q^{\pi}_{obs}(s,act))$ \\
   select new action \textbf{act=actNew} with the max q-value\;
   
   }
  }

 \caption{Q-learning with observations}
 \label{QObs:algorithm}
\end{algorithm}

\begin{algorithm}[t]
\SetAlgoLined
\KwResult{act}

  \eIf{random.uniform(0, 1) $<$ epsilon (the exploration probability)}
  {
   Explore: select a random action \textbf{act}\;
   
   }
   {
   Exploit: select the action \textbf{act} with the max q-value\;
    \If{\textbf{act} leads to adversary}{
   select the action \textbf{act} with the second max q-value\;
   
   }
  }

 \caption{Modified Q-learning}
 \label{Qmodif:algorithm}
\end{algorithm}

\subsection{Formal method}
We have employed formal methods to investigate potential guarantees for the proposed defense mechanisms. We are going to check whether we are able to provide formal guarantees on the agent's behaviour in the environment. To do so, we evaluate the effectiveness of the defense approaches through probabilistic model checking. We verify the agent's reachability properties using the PRISM model checker. We extracted a DTMC after the training process, reflecting the agent's behavior. Since, at each location, the Q-learning algorithm uses the Q-function to determine the agent's next action, the DTMC is constructed by relying on that Q-function to compute the probability of taking a specific action for each transition. We construct the latter for the next $100$ episodes after $2000$ episodes. The nodes of the DTMC are the agent's locations in the environment. The agent's reachability properties are verified on the DTMC. We measure the safety of the agent through the probability of reaching the goal within some particular steps in presence of adversaries.
 
We want to evaluate the agent’s policy augmented with the employed defense mechanisms, trained in an adversarial environment. The agent's policy is verified when acting against the learnable adversary as it is the most challenging adversary in our study. Its reachability properties are formalized onto PRISM in PCTL logic as follows:
\begin{itemize}
    \item  \textbf{P1~:}~$P_{>=1} ~[ G ~ F ~(Agent=g) ]$\\
    \textit{With $100 \%$ probability the agent will always reach its goal.} 
    \item \textbf{P2~:}~$P=?~ [ F_{<=100} ~(Agent=g) ]$\\
    \textit{What is the probability that the agent reaches its goal within 100 steps}
    \item \textbf{P3~:}~$P=?~ [ F~ (Agent=g) ]$\\
    \textit{What is the probability that the agent reaches its goal }
\end{itemize}
where $g$ is the goal's location.

\section{Experiments} \label{section5}
In this section, first we describe our research questions. Then, we present the experimental design for collecting results. Finally, we will explain the conducted experiments to assess the agent's behaviour against adversaries and answer our RQs. The interested reader can refer to the source code \footnote{\url{https://github.com/movingadv/On-Assessing-The-Safety-of-Reinforcement-Learning-algorithms-Using-Formal-Methods}}.

\subsection{Research questions}
We formulate the following research questions to address the proper assessment of the agent's safety:\\
\textbf{(RQ1) What impact do moving adversaries have on the learning of the RL agent?}\\ 
In this research question, we want to determine the level of harm that a moving adversary (learnable or non-learnable) can cause.\\
\textbf{(RQ2) To what extent can PBRS-based defense mechanisms improve the agent’s policy so that the agent can achieve its objective in presence of adversaries?}
This research question aims to evaluate the effectiveness of PBRS as a defense mechanism against the learnable adversary.\\
\textbf{(RQ3) To what extent the direct modification of Q-learning algorithm can improve the learning policy so that the agent can achieve its objective in presence of adversaries?}
This research question aims to evaluate the effectiveness of the modified Q-learning algorithm as a defense mechanism against the learnable adversary. \\
\textbf{(RQ4) What are the formal guarantees that can be provided to the defense mechanisms regarding the agent safety?}\\
In this research question, we provide lower and upper bounds on the agent's reachability properties. The latter assess the effectiveness of the defense strategies.

\subsection{Experimental design}

\begin{figure*}
\begin{subfigure}{0.23\textwidth}
  \centering
   \includegraphics[width=\linewidth]{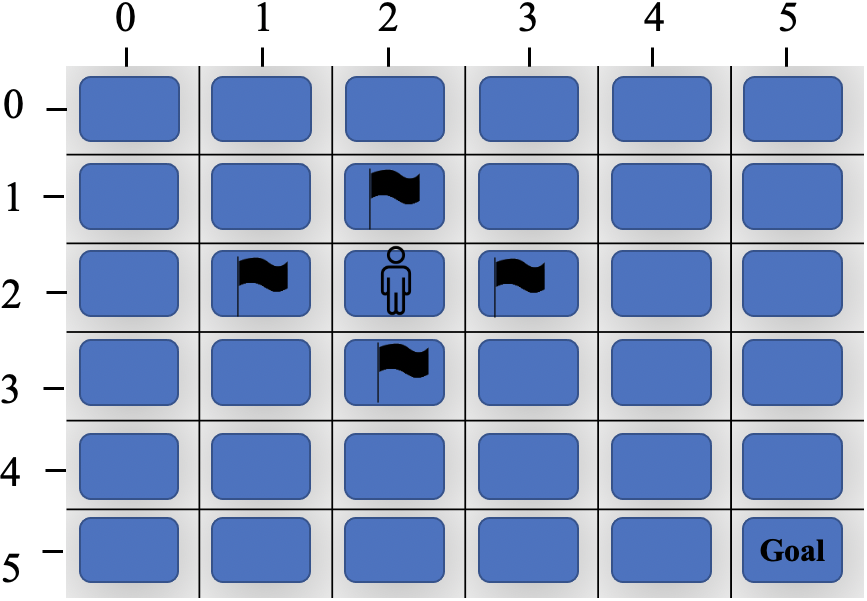}
    \caption{}
    \label{fig:As}
\end{subfigure}
\hfill
\begin{subfigure}{0.23\textwidth}
  \centering
    \includegraphics[width=\linewidth]{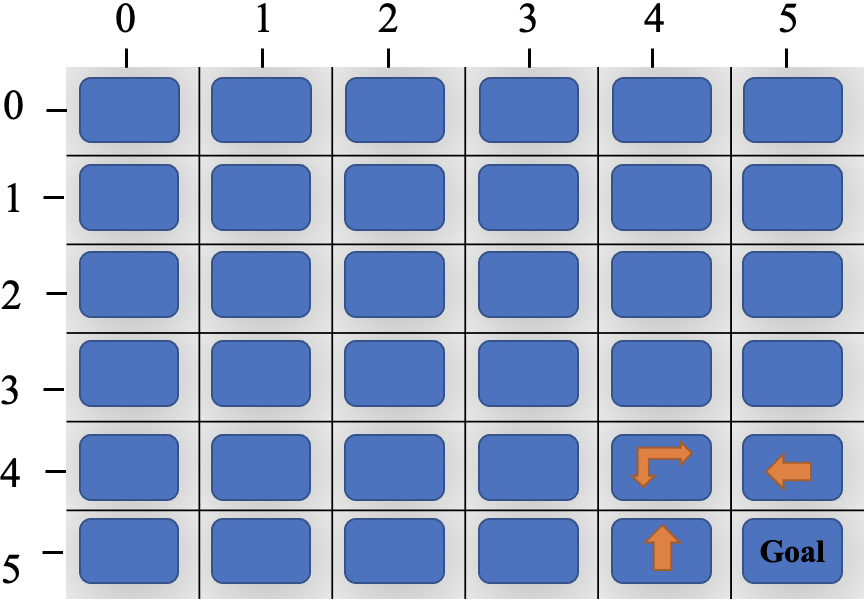}
    \caption{}
    \label{fig:3PatI}
\end{subfigure}
\hfill
\begin{subfigure}{0.23\textwidth}
  \centering
    \includegraphics[width=\linewidth]{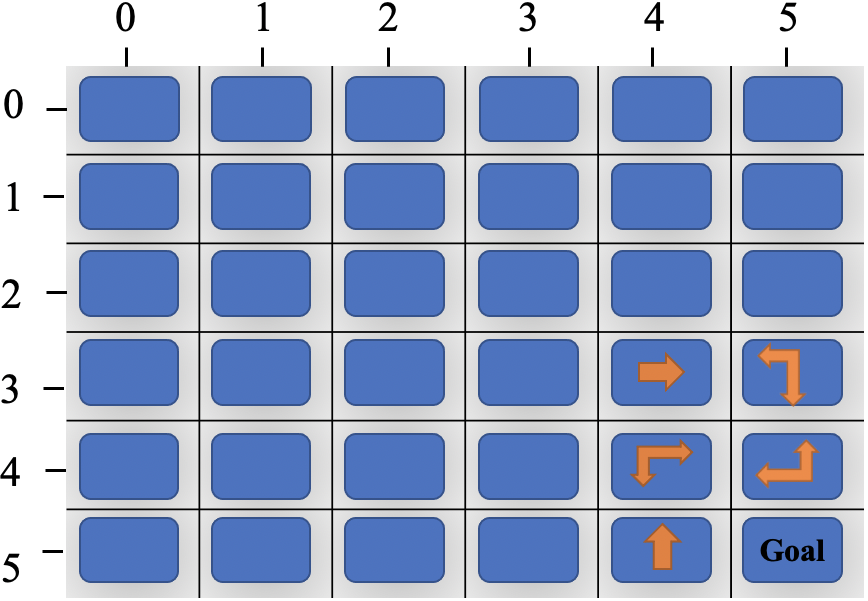}
    \caption{}
    \label{fig:5PatI}
\end{subfigure}
\hfill
\begin{subfigure}{0.22\textwidth}
  \centering
      \includegraphics[width=\textwidth]{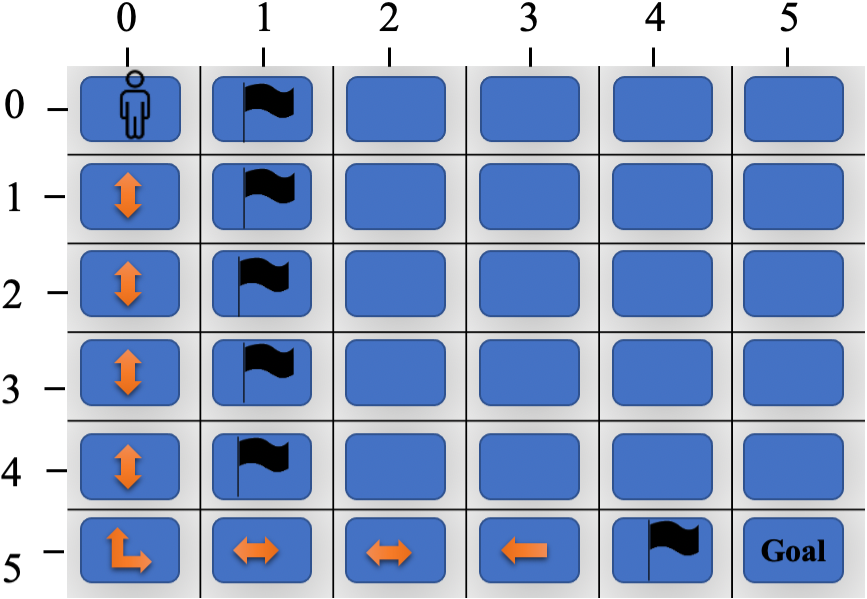}
       \caption{}
   \label{fig:pathQmodif}
\end{subfigure}
\caption{
a) an example of the agent location, its surrounding and the goal location, b) 3-cell patrolling adversary trajectory, c) 5-cell patrolling adversary trajectory, d) The agent's path with the modified Q-learning process employed.}
\label{fig:grids}
\end{figure*}

First of all, we set up an environment for our experiments. We implement a 6 $\times$ 6 grid world where the agent starts from the upper left of the grid ($cell (0,0)$) and has to make its way to the goal's location, i.e., lower right of the grid $cell (5,5)$ . Figure \ref{fig:As} displays a possible location of the agent in the environment, and the black flags represent one of its possible locations at the next step after taking an action. As well as shows the environment and associated cell numbers. The actions correspond to moving to the four compass directions: (\textit{right, up, down, left})


For every non-goal cell, the agent gets a reward of $-1$, and $100$ when reaching the goal cell. The agent will get a $-100$ reward when a collision with the adversary occurs. The objective is to reach the goal using the shortest path and without interruption, which allows the increase of the cumulative reward. We define cumulative reward at episode number $x$, as the summation of reward obtained from episode number $1$ until episode number $x$. Moreover, the agent's behaviour is considered safe if there is no collision with the adversary. The agent is trained via Q-learning algorithm with $\epsilon$-greedy policy. All the reported results conducted with this environment are average of $50$ independent runs and each run lasts for $2000$ episodes. In fact, three events may lead to terminate an episode:
 \begin{enumerate} 
     \item The agent reaches its goal, cell $(5,5)$,
     \item A timeout of $100$ steps per episode is reached. In such a case, no additional reward is given neither to the agent nor the adversary for that episode,
     \item A collision occurred between the agent and the adversary.
 \end{enumerate}
 
For \textbf{(RQ1)}, we have considered that the environment is threatened by three kinds of moving adversaries:
\begin{itemize}
    \item A 3-cell patrolling adversary, patrolling successively back and forth on (5,4), (4,4), (4,5) cells. The adversary moves around these cells at every episode as shown in Figure \ref{fig:3PatI}.
    \item  A 5-cell patrolling adversary, patrolling successively back and forth on (5,4), (4,4), (4,5), (3,5), (3,4) cells. The adversary moves around these cells at every episode as shown  in Figure \ref{fig:5PatI}.
    \item A learning adversary which also uses a Q-learning with $\epsilon$-greedy policy starting at the cell (0,0).   
\end{itemize}


The adversaries’ goal is to create as many collisions as possible with the agent. They get a reward of $100$ when a collision occurs and $0$ otherwise in each episode. By creating collisions with the agent, they demonstrate unsafe behavior throughout training. We have collected the rate of collisions caused by the adversarial policies as well as the cumulative reward obtained by the agent while acting against such adversaries. The rate of collisions represents the average number of collisions in the last $100$ episodes.
 
For \textbf{(RQ2)}, we have collected the collision rate and the cumulative reward obtained by the agent 
when facing the learning adversary as it is the most challenging adversary among those we presented in this paper. Also, the results are collected when the agent's strategy is augmented with the PBRS-based defense mechanism using two different potential functions: the probability of avoiding a collision and the distance to the goal. 
 
For \textbf{(RQ3)}, we have implemented the modified Q-learning algorithm presented in Section \ref{section4}. We run this defense mechanism against the learning adversary and evaluate the agent's behaviour by the collision rate and the cumulative reward. Additionally, for \textbf{(RQ2)} and \textbf{(RQ3)} the results are collected when the agent can observe its surrounding cells. An example of the agent and its surroundings are shown in Figure \ref{fig:grids}(b).
 
For \textbf{(RQ4)}, we have conducted a verification process using PRISM model checker\footnote{\url{https://www.prismmodelchecker.org}}. We verify the agent's behaviour in four scenarios where the agent can observe its neighbouring cells in all scenarios:
\begin{itemize}
    \item \textbf{Scenario 1:} The agent is facing the learning adversary,
    
    \item \textbf{Scenario 2:} The agent is acting against the learning adversary and the PBRS-based defense mechanism is employed using the Manhattan distance to the goal as the potential function,
    
    \item \textbf{Scenario 3:} The agent is facing the learning adversary and the PBRS-based defense mechanism is employed using the probability of avoiding a collision as the potential function,
    
    \item \textbf{Scenario 4:} The agent is acting against the learning adversary and the modified Q-learning algorithm is employed. 
\end{itemize}
 
For simplicity, the agent's states are encoded into PRISM from $0$ to $35$. With $0$ the initial state and $35$ the goal state. 

\subsection{Results of the evaluation}
In the following, we present the obtained results and describe answers to our four research questions.\\

\begin{figure*}
\begin{subfigure}{0.3\textwidth}
  \centering
  
        \includegraphics[width=\linewidth]{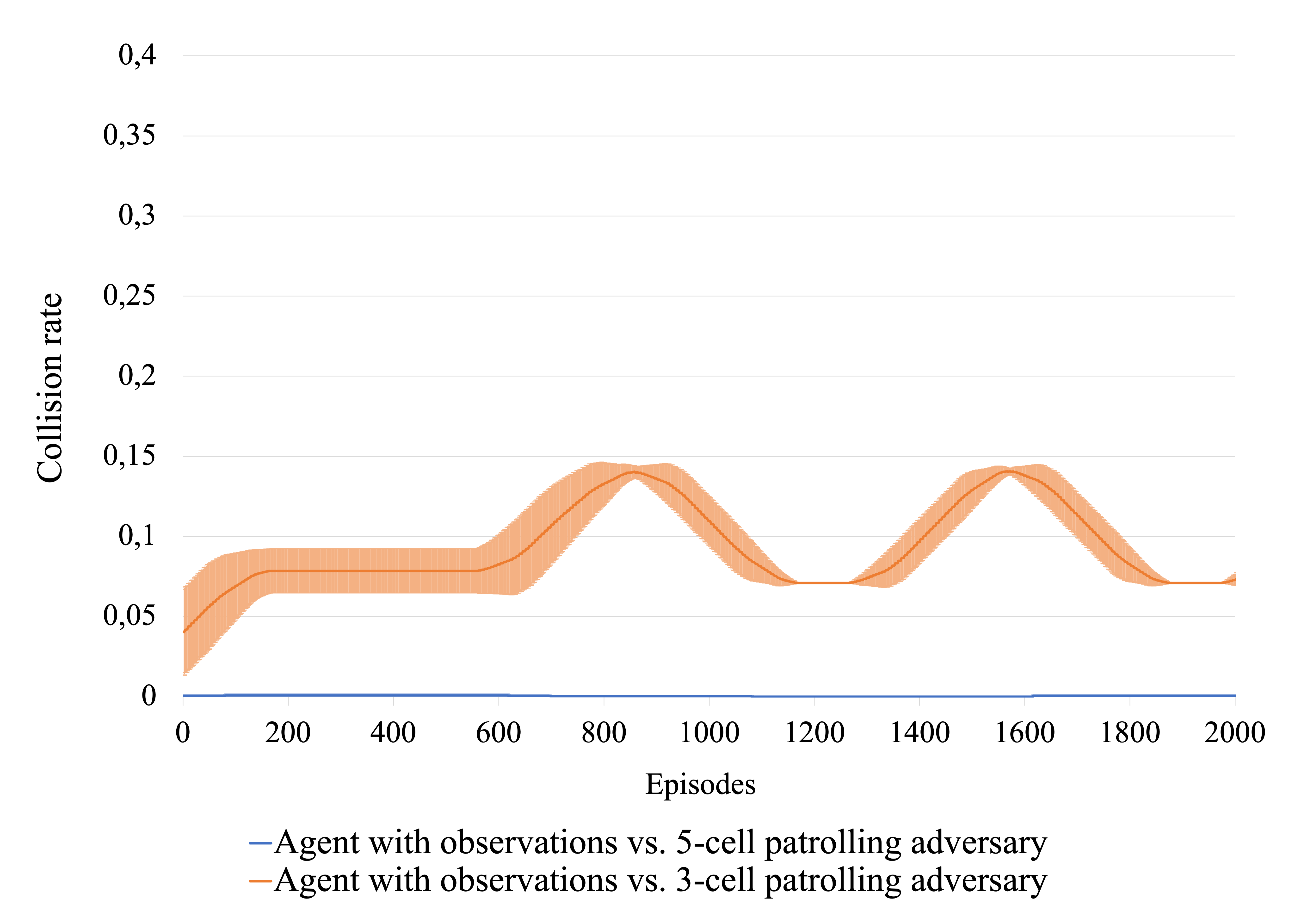}
        \caption{}
        \label{fig:Coll2patAdv}
\end{subfigure}
\hfill
\begin{subfigure}{0.3\textwidth}
  \centering
   \includegraphics[width=\linewidth]{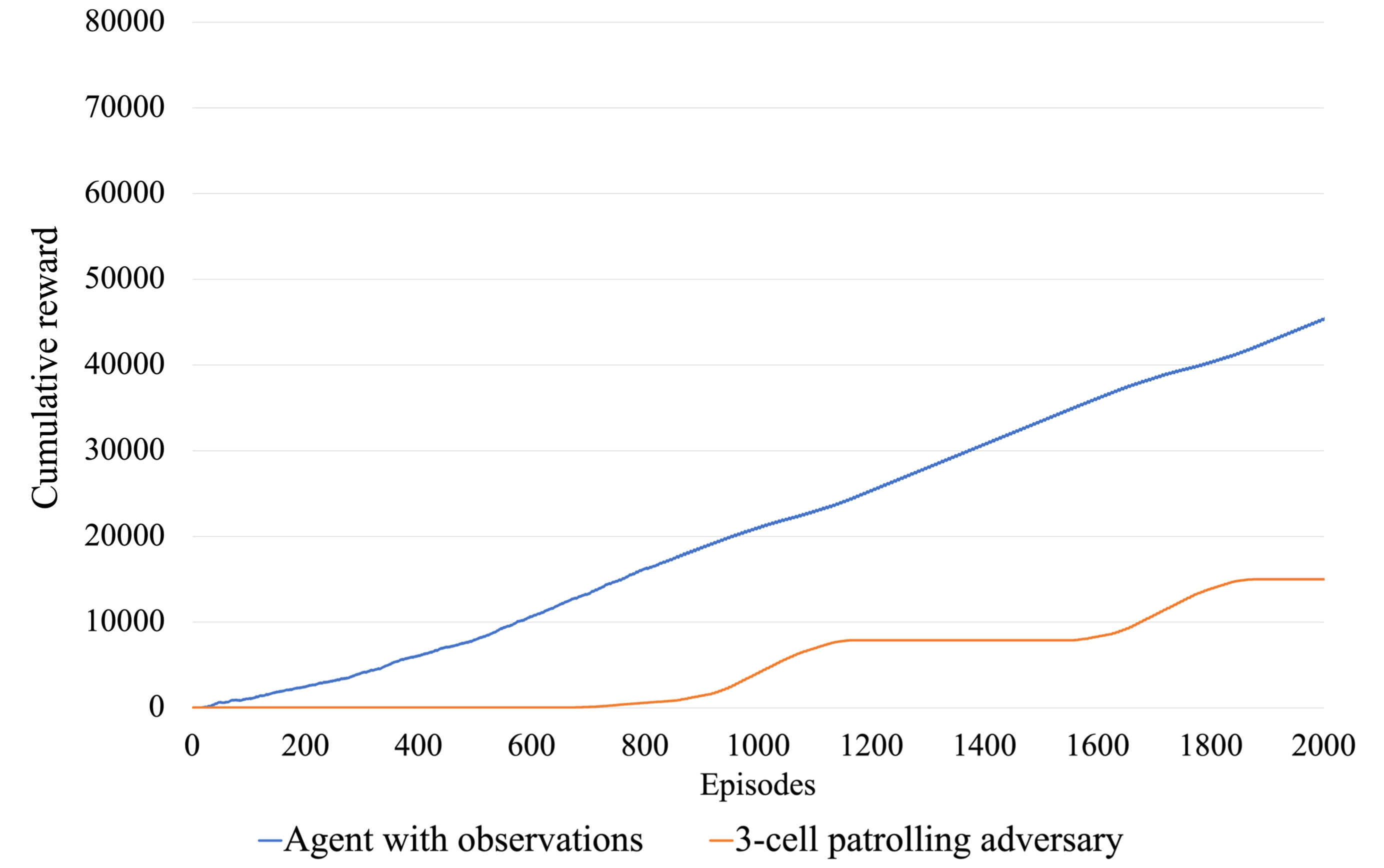}
    \caption{}
    \label{fig:CumReward3cell}
\end{subfigure}
\hfill
\begin{subfigure}{0.3\textwidth}
  \centering
    \includegraphics[width=\linewidth]{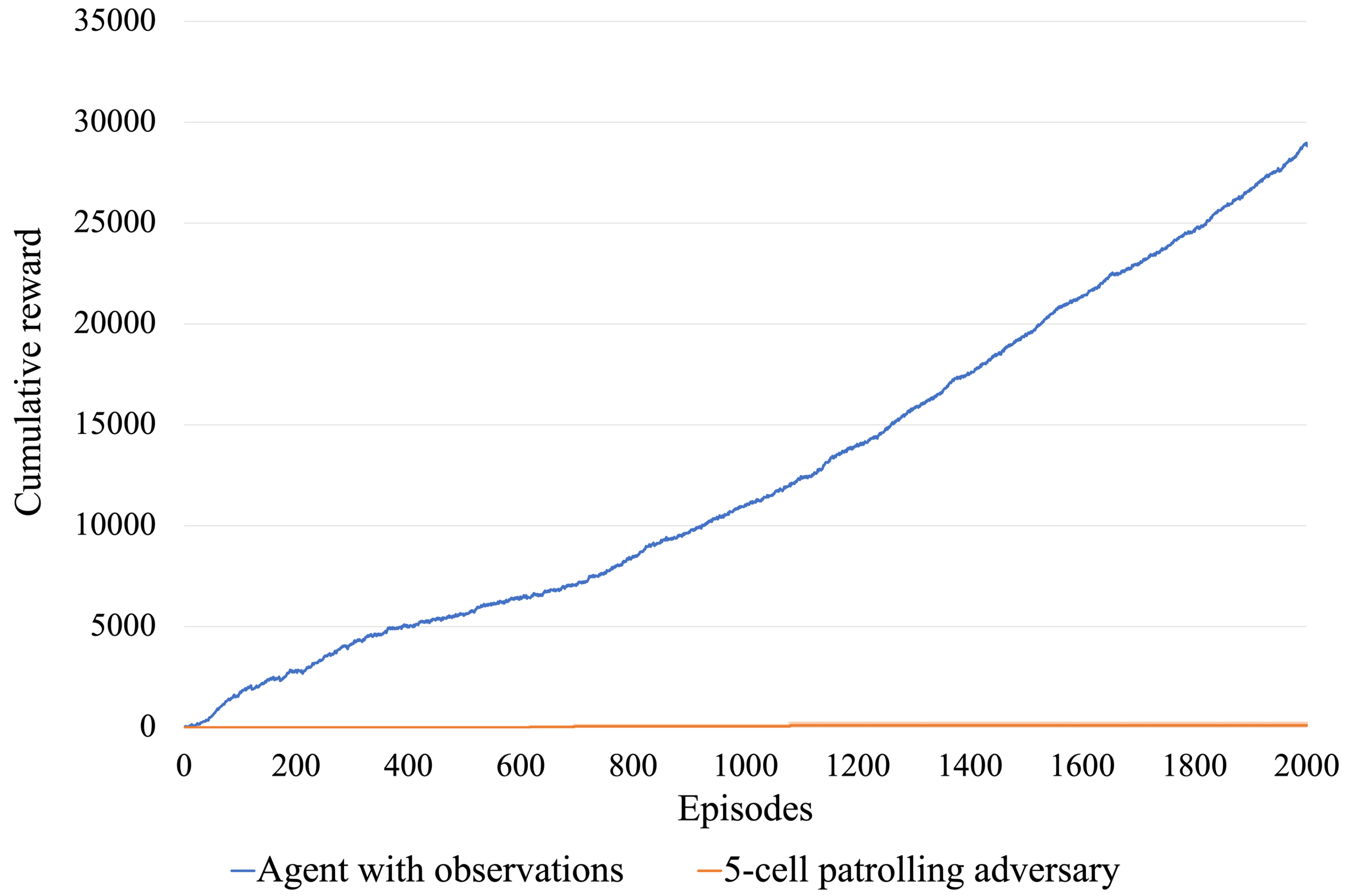}
    \caption{}
    \label{fig:CumReward5Cell}
\end{subfigure}
\caption{a) Collision rate between the agent and patrolling adversaries, b) Cumulative reward of the agent and the 3-cell patrolling adversary, c) Cumulative reward of the agent and the 5-cell patrolling adversary.}
\label{fig:CumReward3cell and CumReward5Cell}
\end{figure*}

\textbf{RQ1: What impact do moving adversaries have on the learning policy of the RL agent?}\\
At episode \#$2000$, the agent faces $86 \%$ collision rate when performing against the learning adversary, $40 \%$ for the 3-cell patrolling adversary and $20 \%$ for the 5-cell patrolling adversary. These results are collected when the agent and the learning adversary do not have any observations. The learning adversary appears to be the most harmful in terms of collision rate created with the agent. When allowing the agent to have observations, we witness almost zero collisions per episode when the agent faces the 5-cell patrolling adversary and less than $20 \%$ collision rate when facing the 3-cell patrolling adversary. The results presented in Figure \ref{fig:Coll2patAdv} report the mean of collision rate for all episodes. 


\begin{table}[t]
\caption{Collision rate between the agent and the adversaries when the agent has observations and no defense mechanisms is applied against these adversaries at Episode \#2000.}
\begin{center}
\resizebox{\linewidth}{!}{
\begin{tabular}{l | c }
\hline
Adversary & Collision rate $\pm$ SD \\
\hline
\textbf{5-cell patrolling adversary} & 0.00 $\pm$ 0.00 \\
\textbf{3-cell patrolling adversary} & 0.07 $\pm$ 0.00 \\
\textbf{Learning adversary with observations} & 0.39 $\pm$ 0.35 \\
\textbf{Learning adversary without observations} & 0.41 $\pm$ 0.15\\ 
\hline
\end{tabular}
}
\label{table:meanSTD3}
\end{center}
\end{table}

Table \ref{table:meanSTD3} includes the value of the collision rate between the agent and the adversary and its standard deviation at the end of experiment (i.e., Episode No. $2000$) for the patrolling adversaries. No defense mechanisms have been applied against the patrolling adversaries due to the low collision rate. These values underlie the effect of the observations on reducing the collision rate, as the agent faces the 3-cell and 5-cell patrolling adversaries. Figures \ref{fig:CumReward3cell} and \ref{fig:CumReward5Cell} show the cumulative reward obtained by the agent and the patrolling adversaries over $50$ runs. Numerical values along with standard deviations are reported on Tables \ref{table:meanSTDCum3cell}.
Larger amounts of the cumulative reward of the agent compared to the patrolling adversaries prove that the agent can reach its goal without interruption despite the presence of patrolling adversaries in the environment. The patrolling adversaries represent less threat to the agent's safety when adding observations to agent’s knowledge. 

\begin{table*}[t]
\caption{Cumulative reward of the agent and the 3-cell, 5-cell patrolling adversaries (cumulative reward, mean reward $\pm$ standard deviation).}
 \centering
\begin{tabular}{c |p{3cm}| p{3.5cm} ||  p{3cm} |  p{2.5cm} }
\hline
 & \textbf{Episode \#500} & \textbf{Episode \#1000} & \textbf{Episode \#1500} & \textbf{Episode \#2000} \\
\hline
\textbf{Agent with observations} & 7892.12, 86.5 $\pm$ 4 & 20979.87, 63.25 $\pm$  69.8 & 33515.62, -77 $\pm$ 0 & 45234.62, 91 $\pm$ 0 \\
\hline
\textbf{3-cell patrolling adversary} & 50, 0 $\pm$0  & 4037.5, 12.5$\pm$ 35.35 & 7900, 0 $\pm$ 0  &  15000, 0 $\pm$ 0 \\
\hline
\hline
\textbf{Agent with observations} & 5607.5, -14 $\pm$ 15.40 & 11043.75, -53.5 $\pm$ 10.13 & 19514.75, -53 $\pm$ 3 & 28974.75, 91 $\pm$ 0 \\
\hline
\textbf{5-cell patrolling adversary} & 0, 0 $\pm$0  & 50, 0 $\pm$ 0 & 100, 0 $\pm$0  &  100, 0 $\pm$0 \\
\hline
\end{tabular}
\label{table:meanSTDCum3cell}
\end{table*}

\begin{figure*}

\begin{subfigure}{0.42\textwidth}
  \centering
    \includegraphics[width=\linewidth]{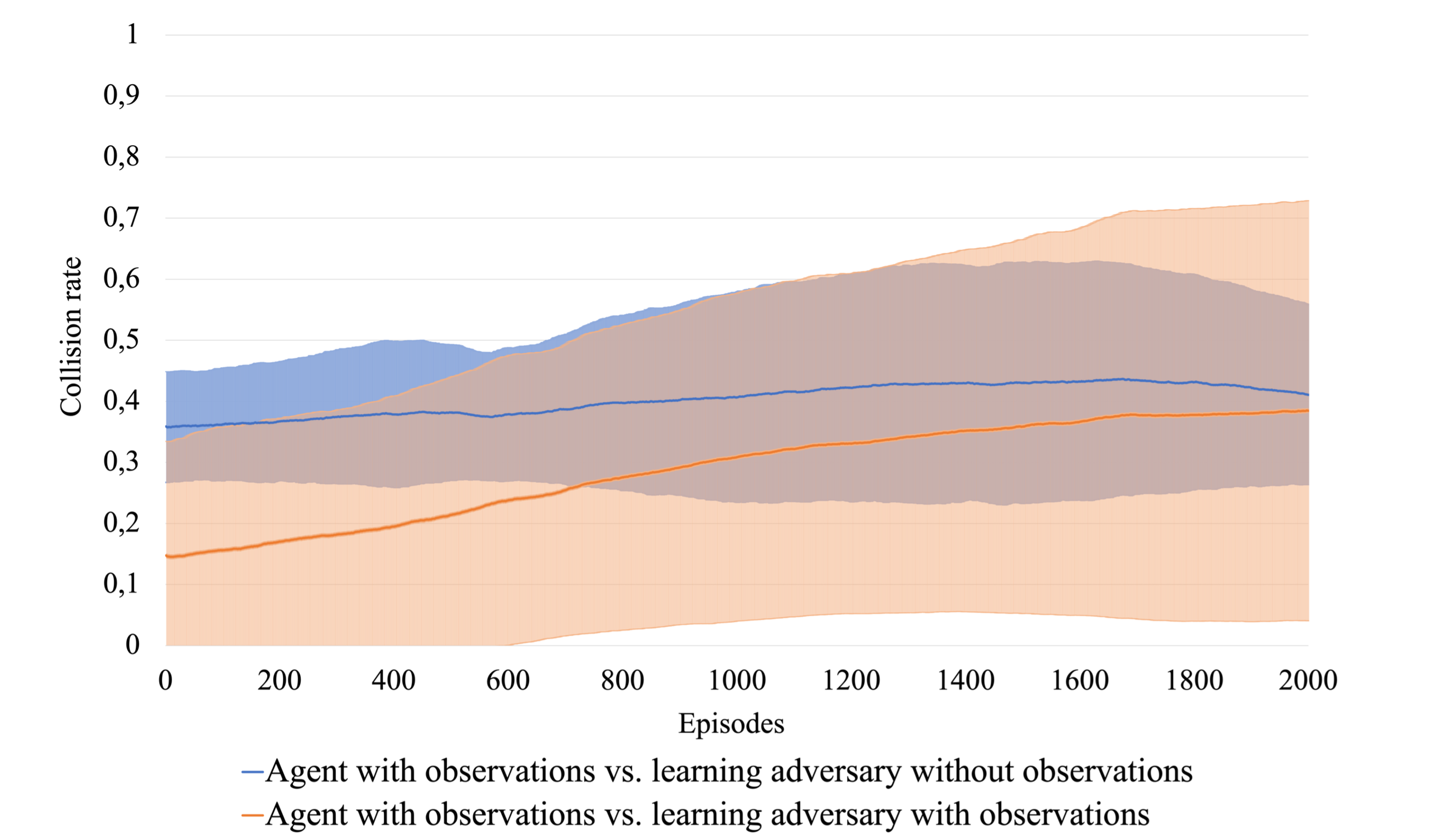}
    \caption{}
    \label{fig:Coll2LearnAdv}
\end{subfigure}
\hfill
\begin{subfigure}{0.42\textwidth}
  \centering
  \includegraphics[width=\linewidth]{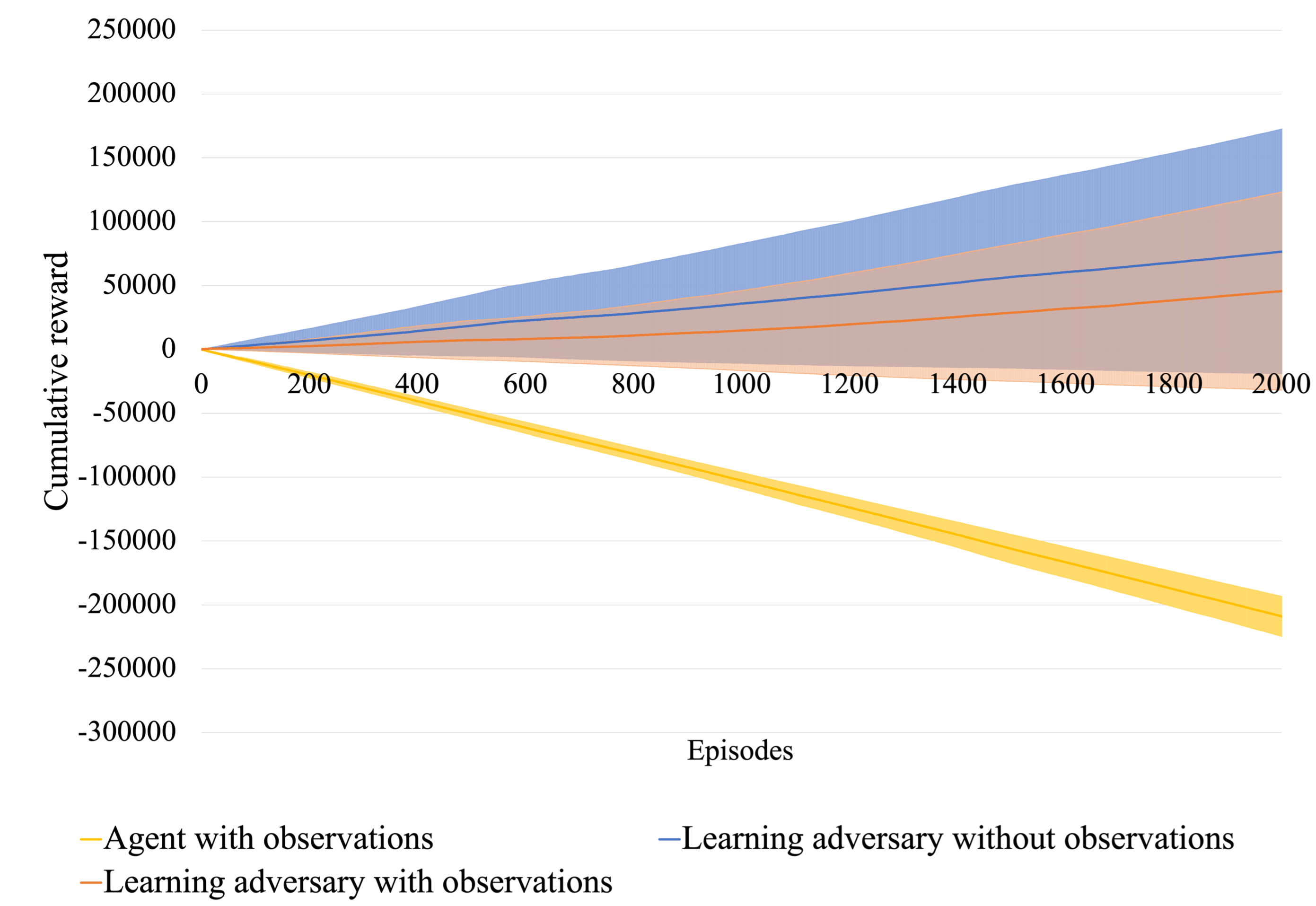} 
\caption{}
\label{fig:MergeLearn+O-O}
\end{subfigure}
\caption{a) Collision rate between the agent and the adversary with and without observation, b) Cumulative reward of the agent and the learning adversary with and without observation.}
\label{fig:two graphs1}
\end{figure*}

Nevertheless, the agent's policy with observations is not that successful in presence of the learning adversary. The learning adversary learns to predict the agent's moves which result in high collision rate, negative rewards earned and then a decrease of the cumulative reward earned by the agent in comparison to patrolling adversaries. We did two experiments regarding the behaviour of the learning adversary to find out if a learning adversary with observations is more harmful. Figure \ref{fig:Coll2LearnAdv} reports at each episode the collision rate between the agent and the learning adversary along with the corresponding standard deviation, when the agent and the learning adversary have observations and when they do not. Figure \ref{fig:MergeLearn+O-O} reports the cumulative reward with the corresponding standard deviation for these experiments. Table \ref{table:meanSTD3} and Table \ref{table:meanSTD} report the collision rate between the agent and the adversary, the cumulative reward, and their standard deviation respectively at episode \#$2000$. Although it has observations, the learning adversary does not appear to be more harmful. Its observations do not affect the agent's behaviour, since the latter still sees its surroundings and can manage to avoid the adversary. So, for the rest of our experiments, we focus on a learning adversary without observations.
 

\begin{table}
\caption{Comparison between the environment configurations at Episode \#2000. The agent can observe its surroundings.}
\centering
\resizebox{\linewidth}{!}{
\begin{tabular}{p{2.6cm} || c || c }
\hline
 Configuration & Cumulative reward & Mean reward $\pm$ SD \\
\hline
\textbf{Agent} & -208902.67 & -101.50 $\pm$ 0.86 \\
\hline
\textbf{Learning adversary without observations} & 76557.14 & 28.57 $\pm$ 48.79 \\
\hline
\textbf{Learning adversary with observations} & 45625.00 & 25.00 $\pm$ 46.30\\ 
\hline
\end{tabular}
}
\label{table:meanSTD}
\end{table}

\begin{figure}
  \centering
        \includegraphics[width=0.8\linewidth]{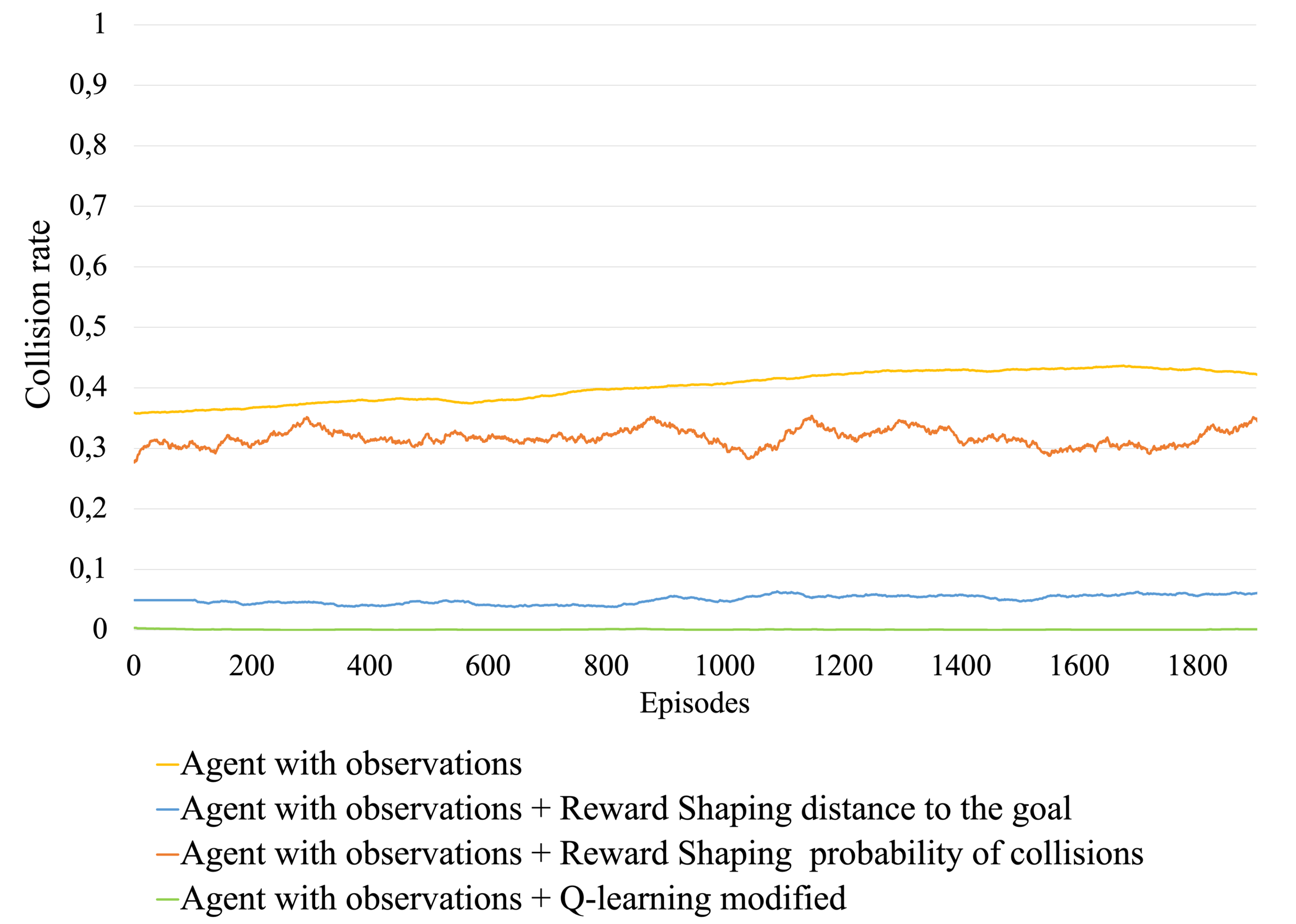}
        \caption{Comparison between collision rates after applying the defense mechanisms against learning adversary.}
        \label{fig:Collplus3Def}
\end{figure}
\begin{figure*}
\begin{subfigure}{0.42\textwidth}
  \centering
  \includegraphics[width=\textwidth]{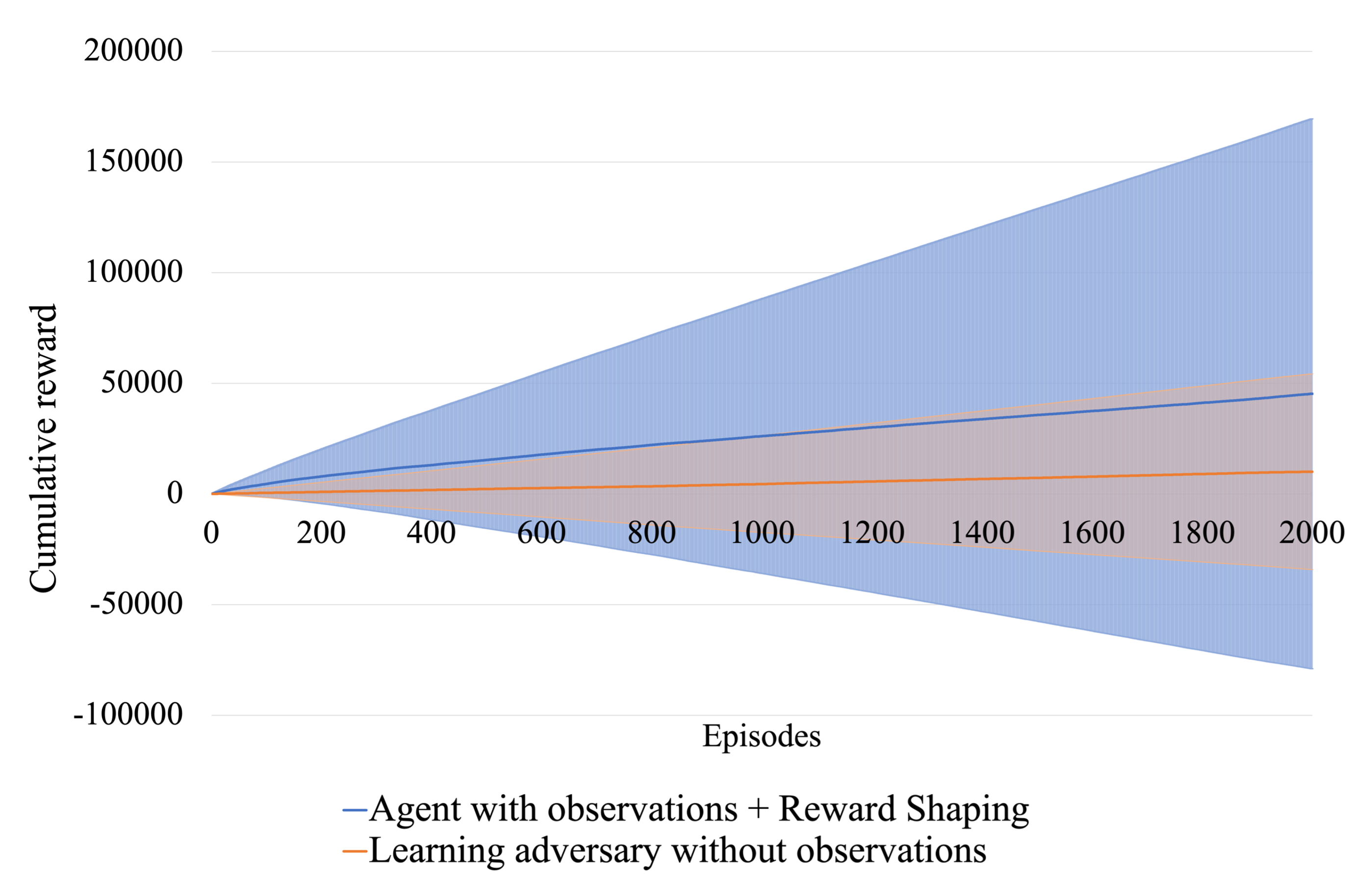} 
   \caption{}
  \label{fig:CumRewardlearnWithoutDef}
\end{subfigure}
\hfill
\begin{subfigure}{0.42\textwidth}
  \centering
  \includegraphics[width=\textwidth]{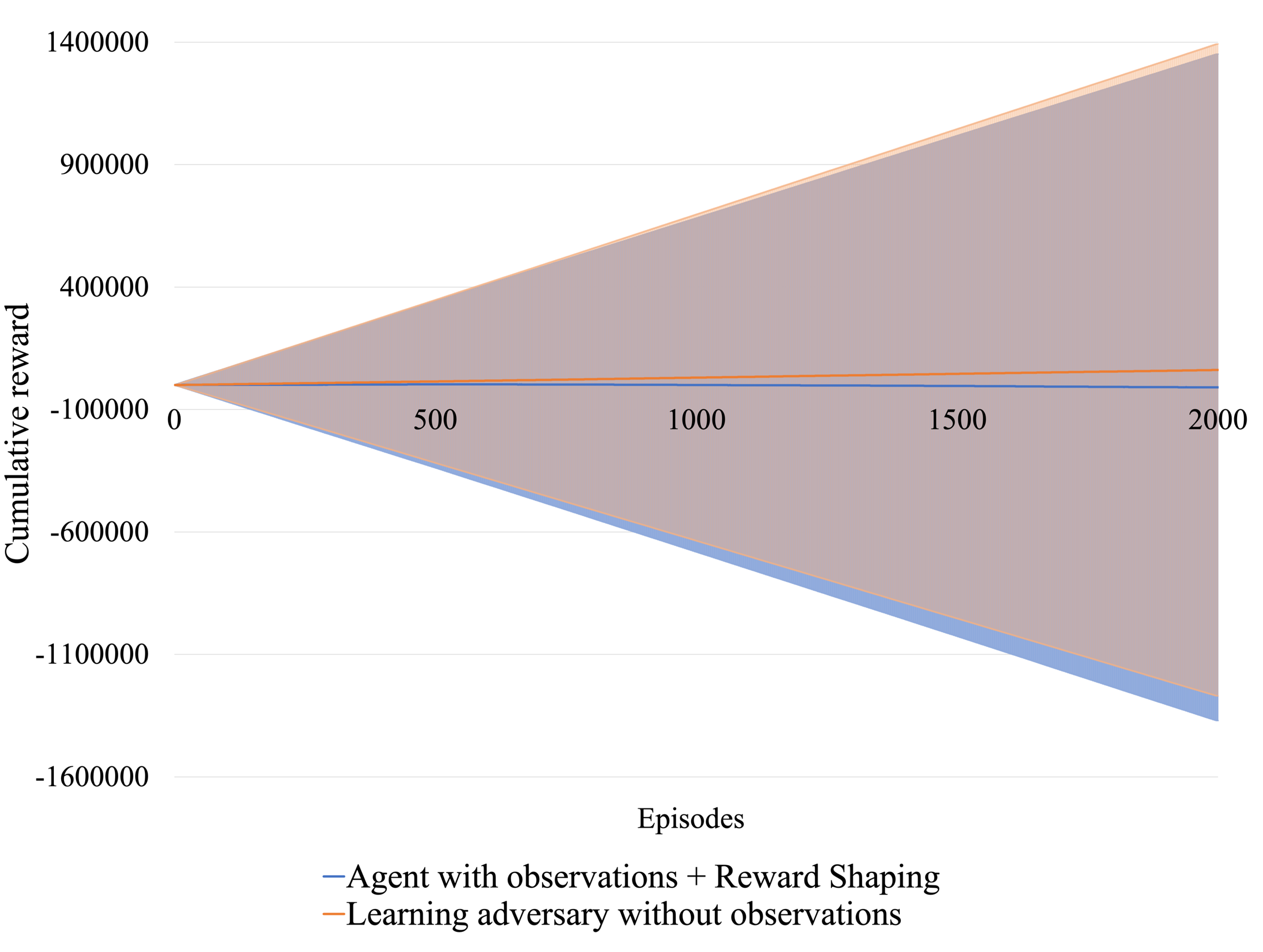}  
   \caption{}
  \label{fig:CumRewardlearnWithoutDef2}
\end{subfigure}

\caption{a) The potential function is the distance to the goal, b) The potential function is the probability of avoiding a collision.}

\label{fig:two graphs2}
\end{figure*}

\textbf{RQ2: To what extent can PBRS-based defense mechanisms improve the agent’s policy so that the agent can achieve its objective in presence of adversaries?}\\
The collision rate after applying both PBRS-based defense mechanisms are shown in Figure \ref{fig:Collplus3Def} along with standard deviation. As in Table \ref{table:meanSTD5}, the value of the collision rate obtained by the agent equipped with defense mechanisms at episode \#$2000$, shows that these mechanisms were able to reduce collisions between the agent and the learning adversary. However, PBRS with the probability of avoiding a collision as a potential function did not contribute to improving the learning as the cumulative reward still decreases during training. Figure \ref{fig:CumRewardlearnWithoutDef2} reports this decrease in cumulative reward and the corresponding standard deviation. The reason is that the location of the learning adversary changes at each step which reduces the accuracy of the probability of avoiding a collision at the next step. 


\begin{table}[tbp]
\caption{Collision rate collected of the environment configurations after applying the defense mechanisms (collision rate $\pm$ standard deviation).}
\centering
\resizebox{0.9\linewidth}{!}{
\begin{tabular}{l | c }
\hline
Defense mechanism & Episode \#2000\\
\hline
\textbf{Modified Q-learning} & 0.00 $\pm$ 0.14 \\
\textbf{PBRS (distance to goal)} & 0.05 $\pm$ 0.14 \\
\textbf{PBRS (probability of collisions)} & 0.30 $\pm$ 0.30 \\
\hline
\end{tabular}
}
\label{table:meanSTD5}
\end{table}

\begin{figure}
  \centering
        \includegraphics[width=0.8\linewidth]{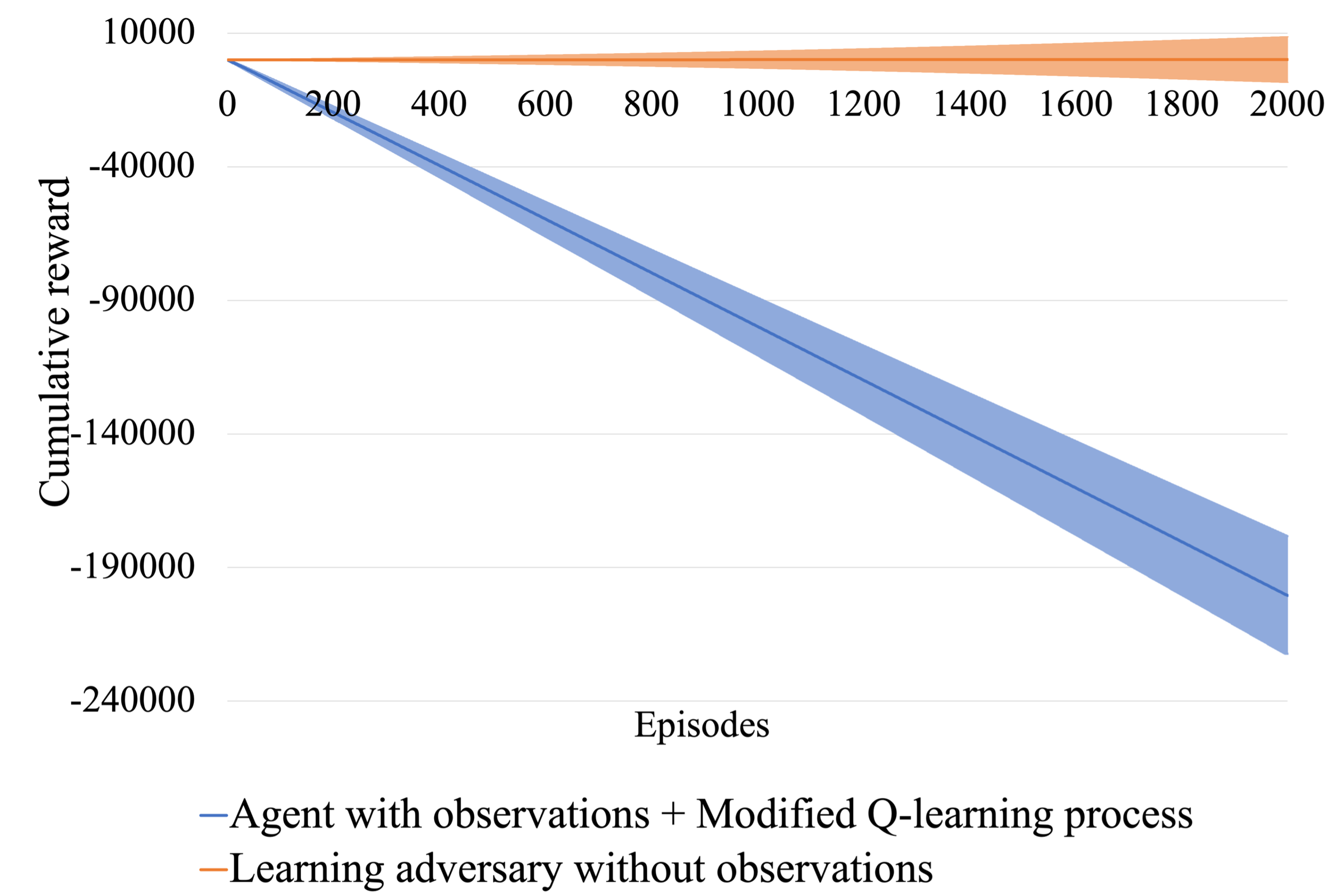}
        \caption{Cumulative reward earned by the adversary and the agent with the modified Q-learning process}
        \label{fig:CumrewardQmodif}
\end{figure}
PBRS with the distance of the agent to the goal as a potential function presents better performances than the previous one. The agent is capable of reaching the goal while lowering its collision rate with the learning adversary and increases its cumulative reward. Figure \ref{fig:CumRewardlearnWithoutDef} reports the cumulative reward obtained by the agent with the corresponding standard deviation when the distance to the goal location is employed as a potential function. Table \ref{table:meanSTD2} reports, at episode \#$2000$, a better cumulative reward when using PBRS with the distance of the agent to the goal as a potential function in comparison to the probability of avoiding a collision.\\

\textbf{RQ3: To what extent the direct modification of Q-learning algorithm can improve the learning policy so that the agent can achieve its objective in presence of adversaries?}\\
The collision rate collected after applying the modified q-learning process is shown in Figure \ref{fig:Collplus3Def}. We have reported a low collision rate (Table \ref{table:meanSTD2}) but a decreasing cumulative reward, as in Figure \ref{fig:CumrewardQmodif}. The modified Q-learning algorithm appears to be conservative as the priority for the agent is to avoid the learning adversary at the expense of not reaching the goal. Figure \ref{fig:pathQmodif} shows an example of a path taken by the agent when observing the adversary in its surroundings. The agent was on its way to the goal location, but since it has to avoid the adversary, the chosen path does not include the goal eventually. Also, we observe in some cases that the standard deviation is greater than the mean. This is due to the disparity between the reward obtained by both the adversaries and the agent.\\

\begin{table*}
\caption{Comparison between different environment configurations.}
\centering
\begin{tabular}{l | c | c}
\hline
Configuration & Cumulative reward & Mean reward $\pm$ SD \\
\hline
\hline
\textbf{Agent with observations + PBRS (distance to goal)} & 45240.10 & 29.02 $\pm$ 62.43 \\
\textbf{Learning adversary without observations} & 10050 & 2.00 $\pm$ 22.73 \\
\hline
\hline
\textbf{Agent with observations + PBRS (probability of collision)} & -126488.30 & 28.73 $\pm$ 681.28 \\
\textbf{Learning adversary without observations} & 63330 & 10.00 $\pm$ 666.67 \\
\hline
\hline
\textbf{Agent with observations + Modified Q-learning} & -200476.54 & -101.00 $\pm$ 12.22 \\
\textbf{Learning adversary without observations} & 142 & 0.00 $\pm$ 6.72 \\
\hline
\end{tabular}
\label{table:meanSTD2}
\end{table*}


\textbf{RQ4: What are the formal guarantees that can be provided to the defense mechanisms regarding the agent safety?}\\
We choose reachability analysis to verify the capability of the agent to reach its goal without facing an obstacle. The results of the verification are formal guarantees regarding the safety of the trained agent against the moving adversaries. The verification process consists of checking for each property the paths that satisfy the following scenarios. We have verified properties \textbf{P1, P2} and \textbf{P3} on \textbf{Scenario 1, Scenario 2, Scenario 3} and \textbf{Scenario 4}. We have reported the results in Table \ref{table:resultV}. The decrease of the cumulative reward on \textbf{Scenario 1, Scenario 3} and \textbf{Scenario 4} explained well the results of the verification process. The verification process provides with no doubt that the environment remains unsafe after applying the defense mechanisms to these scenarios. On the contrary, in \textbf{Scenario 2} we can see that the applied defense mechanism helps the agent to reach its goal with a minimum probability of $10 \%$. More explanatory, within $100$ steps the agent has $10 \%$ chance to complete its task. Its chance increases when the timeout happens more frequently as well. 

\begin{table}[t]
\caption{Results of probabilistic model checking.}
\centering
\begin{tabular}{p{2cm} || p{1.5cm} ||  p{1.5cm} ||  p{1.5cm} }
\hline
 & \textbf{P1} & \textbf{P2} & \textbf{P3} \\ [1ex] 
\hline\hline
\textbf{Scenario 1} &False&0&0\\
\textbf{Scenario 2}&True&0.1&1\\
\textbf{Scenario 3}&False&0&0\\
\textbf{Scenario 4} & False&0&0\\[1ex]
\hline
\end{tabular}
\label{table:resultV}
\end{table}

\section{Related work} \label{section2}
An RL process is usually formulated as a Markov Decision Process (MDP), one can see those parts as opportunities to attack an adversary. Hence, the ability to attack RL applications sometimes comes from the interdependence of different actions taken by the agent \cite{ilahi2020challenges}. Initially, the agent is in an exploration phase, making it difficult to differentiate between legitimate and illegitimate actions. Also, the policy followed by the agent to achieve the objective may be deterministic or stochastic, having loopholes that can be used by an attacker. Finally, the manipulation of the environment modifies the observations and the actions taken by the agent. If those manipulations lead to unsafe states, the final reward obtained by the agent can be corrupted. By compromising the policy, the overall performance of the agent can be degraded. Lin et al. \cite{lin2017detecting} showed the presence of adversarial samples on Neural Networks policies, by leveraging a prediction frame module. Moreover, RL agents face misspecifications \cite{mankowitz2019robust} due to setting changes between the training environment and the test environment. These modelling errors have a negative impact on the agent's policy and future rewards \cite{morimoto2005robust}. Adversarial samples can also be detected by introducing agents with adversarial policies in the environment. Wang et al \cite{wang2020reinforcement} propose techniques to uncover flaws in the training of the agent, by introducing an attacker which leads the agent to failure by perturbing the rewards observed. Pinto et al. \cite{pinto2017robust}, use a zero-sum games configuration to model failures inside the environment. Antagonists players described a policy that is harmful to the agent. Our work is similar to these previous works since we introduced an adversary which negatively challenges the agent in achieving its goal.
 
Several works have investigated mechanisms to help agents face perturbations that might occur during the training. Augmenting the reward distribution of the agent can improve its learning policy when doing it in an appropriate way. Some other works on reward shaping to improve the learning policy include automatic reward shaping approaches \cite{grzes2008learning} \cite{marthi2007automatic}, multi-agent reward shaping \cite{devlin2011theoretical} \cite{sun2018designing}, and some novel approaches such as belief reward shaping \cite{marom2018belief}, ethics shaping \cite{wu2018low}, and reward shaping via meta-learning \cite{zou2019reward}. In our work, we implement PBRS as our defense strategy to face adversarial policy. Our potential function is partially inspired by the work done in \cite{ng1999policy} and adapted to the model-free RL. The authors propose a potential function, the manhattan distance between the agent location and its goal. They show that it is an effective approach that can be implemented by researchers when trying to improve their RL systems.
 
Adversarial training which consists of retraining a machine learning model with adversarial examples to increase the robustness of the model against adversarial examples can be implemented as defense mechanisms. Researchers in \cite{kos2017delving}, \cite{pattanaik2018robust}, \cite{han2018reinforcement}, \cite{behzadan2017whatever}, proposed to re-train the model with perturbations generated by adding noises to states and rewards during the training. In the game-theoretic approaches the attacker is considered to be competing in a game with the agent, they adjust their choices based on the payoff during the training. Ogunmolu et al. \cite{ogunmolu2018minimax} propose a minimax iterative dynamic game framework for designing robust policies in the presence of adversarial inputs. Pinto et al. \cite{pinto2017robust} modelled interactions between both the agent and the attacker as a zero-sum game. The agent improves its policy by trying to win the attacker. In presence of modelling errors or misspecifications of training parameters, one can implement a robust method that incites the agent to learn only the optimal policy \cite{mankowitz2019robust}, \cite{morimoto2005robust}. Li et al. \cite{li2018training} propose a robust controller that allows the agent to face model uncertainties and external disturbances. In comparison to adversarial training, the agent is retrained in presence of adversaries, but with additional knowledge about them. This additional information is incorporated in its training algorithm.
 
The above studies have contributed to improving the RL agent's learning. However, they suffer from the lack of guarantees on the safety of the agent which reduces their applicability. In this paper, we propose to implement formal methods that have provided strong guarantees on the expected behaviour of systems including software systems and critical systems. Some researchers have used formal methods to improve the agent's policy. They considered the MDP of the environment to build a verification strategy. Mason et al. \cite{mason2017assured} generate a set of policies and force the agent to learn only policies that satisfy a set of predefined constraints. Probabilistic model checking has also been applied to verify the satisfiability of safety requirements \cite{hasanbeig2018logically}, \cite{pathak2018verification}. Li et al. \cite{li2019formal} formulate safety requirements into temporal logic formulas which they use to construct a finite-state automaton. The automaton then synthesizes a safe optimal controller which can notify when the system enters into traps or executes tasks that are impossible to be completed.  Alshiekh et al. \cite{alshiekh2018safe} propose a method for the RL agent to learn optimal policies on the MDP of the environment while enforcing temporal logic properties. These works assume the model of the environment is known. In our work the agent is trained through Q-learning, a model-free RL algorithm and the policy is represented by a state action-value function (Q function). In \cite{bouton2019reinforcement}, the authors provide probability guarantees by computing probabilistic reachability properties using a value iteration algorithm. Our work has similarities with this work as we also provide probabilistic reachability properties for a set of states, but we rely on a model checker to compute those probabilities. Moreover, we are able to provide formal guarantees after the verification process on a set of desirable properties. 
 
\section{Conclusion and Discussions} \label{section6}
In this paper, we have presented an approach to assess the safety of an RL agent, using formal methods. The proposed approach consists of designing moving adversaries that harm the agent's behaviour. This leads to implementing some defense mechanisms to strengthen the agent's policy. The experiments show that the defense mechanisms improve the learning in terms of reducing the collision rate with the adversaries. We employ the probabilistic model checking on four scenarios, as mentioned on the Experiments section \ref{section5},to provide provable safety guarantees of the agent's behaviour. During the verification process, we consider simple reachability properties to provide provable guarantees for the safety of the agent’s behaviour. A learnable adversary, which is the most harmful adversary that we design, demonstrates resilience to our defense mechanisms in comparison to the non-learnable adversaries. This is due to its high capability to threaten learning. Nevertheless, according to our formal verification neither of evaluated scenarios is safe for the agent. For future work, we plan to investigate more effective defense mechanisms against the learnable adversary. The main idea is to deceive the adversary so that she can not predict the agent’s behaviour and make collisions. Moreover, we plan to examine the proposed adversaries and defense mechanisms in more complex environments.


\balance
\bibliography{references}
\bibliographystyle{ieeetr}
\end{document}